\documentclass[journal]{IEEEtran}

\usepackage{graphicx}
\usepackage[utf8]{inputenc}
\usepackage{graphics} 
\usepackage{epsfig} 
\usepackage{mathptmx} 
\usepackage{amsmath} 
\usepackage{amssymb}  
\usepackage{cite}
\usepackage{physics}
\usepackage{multicol}
\usepackage{mathtools, cuted}
\usepackage[cmintegrals]{newtxmath}
\usepackage{epstopdf}
\usepackage{graphics} 
\usepackage{epsfig} 

\usepackage{mathptmx} 
\usepackage{amsmath} 
\usepackage{amssymb}  
\usepackage{cite}
\usepackage{multicol}
\usepackage{mathtools, cuted}
\usepackage[cmintegrals]{newtxmath}
\usepackage{mathtools}

\usepackage{mathptmx}
\usepackage{helvet}
\usepackage{courier}
\usepackage{subfigure}
\usepackage{type1cm}
\usepackage{titlesec}
\usepackage{epsfig} 
\usepackage{mathptmx} 
\usepackage{amsmath} 
\usepackage{amssymb}  
\usepackage{cite}
\usepackage{multicol}
\usepackage{mathtools, cuted}
\usepackage[cmintegrals]{newtxmath}
\usepackage{makeidx}         
\usepackage{algorithm}
\usepackage{algpseudocode}

\usepackage{multicol}        
\usepackage[bottom]{footmisc}
\usepackage{caption}
\usepackage{nomencl}
\usepackage[usenames, dvipsnames]{color}
\usepackage{url}
\usepackage{siunitx}

\usepackage{multicol}        
\usepackage[bottom]{footmisc}
\usepackage{caption}
\usepackage{titlesec}
\usepackage{nomencl}
\usepackage[usenames, dvipsnames]{color}
\usepackage{multirow}
\usepackage{adjustbox}
\usepackage{amsthm}
 
\theoremstyle{definition}
\newtheorem{definition}{Definition}

\newtheorem{theorem}{Theorem}

\theoremstyle{remark}
\newtheorem{remark}{Remark}

\newtheorem{assumption}{Assumption}
\ifCLASSINFOpdf
\else
\fi
\begin{document}
%
\title{Quadcopter Team Configurable Motion Guided by a Quadruped}
%
%
%

\author{Mohammad Ghufran, Sourish Tetakayala, Jack Hughes, Aron Wilson, and Hossein Rastgoftar
\thanks{Authors are with the Aerospace and Mechanical Engineering Department, University of Arizona, Tucson, Arizona, USA, {\tt\small \{ghufran1942, sourish, jath03, aronmathias, hrastgoftar\}@arizona.edu}
}
}

\maketitle

\begin{abstract}
The paper focuses on modeling and experimental evaluation of a quadcopter team configurable coordination guided by a single quadruped robot.
We consider the quadcopter team as particles of a two-dimensional deformable body and propose a two-dimensional affine transformation model for safe and collision-free configurable coordination of this heterogeneous robotic system. The proposed affine transformation is decomposed into translation, that is specified by the quadruped global position, and configurable motion of the quadcopters,  which is determined by a nonsingular Jacobian matrix so that the  quadcopter team can safely navigate a constrained environment while avoiding collision. We propose two methods to experimentally evaluate the proposed heterogeneous robot coordination model. The first method measures real positions of quadcopters, quadruped, and environmental objects all with respect to the global coordinate system. On the other hand, the second method measures position with respect to the local coordinate system fixed on the dog robot which in turn enables safe planning the Jacobian matrix of the quadcopter team while the world is virtually approached the robotic system.
\end{abstract}


%
\IEEEpeerreviewmaketitle

\section{INTRODUCTION}

Given the increasing occurrence and severity of natural disasters and emergencies, it is crucial to utilize sophisticated technical solutions to improve the efficiency of search and rescue operations. With the potential to save more lives, shorten reaction times, and lower the risks exposed to human rescuers and first responders, the deployment of robotics in these vital missions represents a significant advancement.  
Researchers have extensively investigated the drone usability for the save and rescue operations have been extensively the past \cite{9321707, pereira2021optimal, rastgoftar2018cooperative}. Although, drones can significantly improve data aerial surveillance and imaging capabilities, they may not be able to effectively search the ground targets. Motivated by this limitation, this work aims to develop  model and experimentally test configurable motion of a heterogeneous robotic system that  consists of the team of quadcopters guided by a single quadruped robot.



\begin{figure}[ht]
    \centering
    \includegraphics[width=\linewidth]{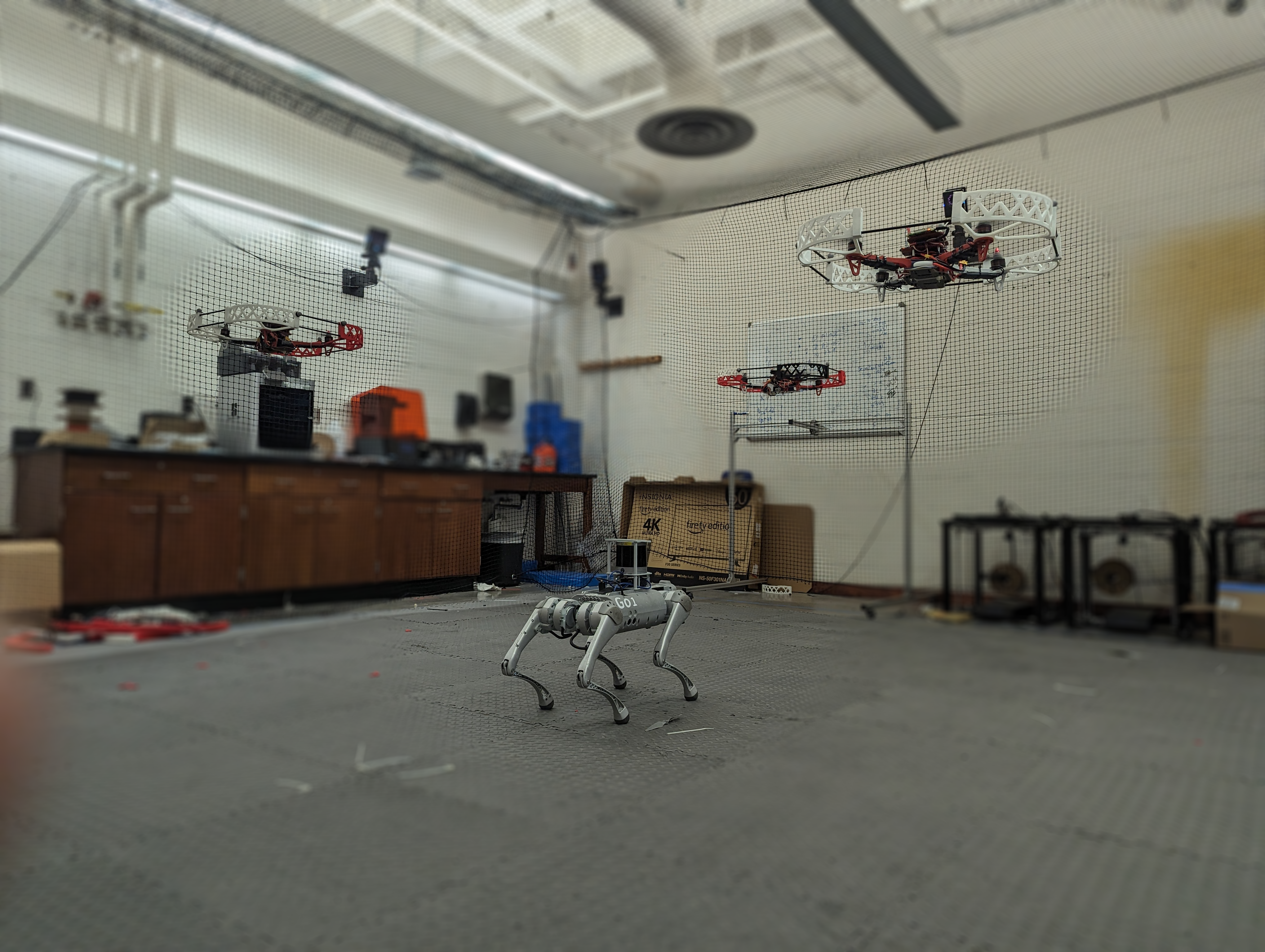}
    \caption{System overview}
    \label{systemoverview0}
\end{figure}

\subsection{Related Work}

 In the ever-evolving field of collaborative robotics, the focus of innovation lies in creating diverse teams of robots that can navigate through complicated settings filled with obstacles. The examination consolidates the influential research presented in \cite{9220149, a72eb80853e4446ba0853f34b53821d5, 8794246, Carney_2022, BAI2018145, 4745874, 6095172, 10.1007/978-3-030-43890-6_22}, to establish a thorough basis for coordinating a team of quadcopters under the guidance of a quadruped robot. The convergence of knowledge highlights the potential advantages and difficulties connected with this navigation method, creating the conceptual foundation of our research.

The studies described in \cite{10.1007/978-3-030-43890-6_22, a72eb80853e4446ba0853f34b53821d5} provide insights into the synergistic capabilities of mixed robotic teams operating in difficult and GPS-denied environments. This research highlights the need to combine aerial and terrestrial robots to improve object recognition and environmental mapping. 

Advanced techniques for decentralized formation coordination  is discussed in \cite{8794246}, \cite{rastgoftar2022spatio},  and \cite{Carney_2022}. These strategies highlight the importance of preserving formation integrity and strong communication connections, especially under difficult conditions. Utilizing graph-theoretic techniques for decentralized control is crucial to our methodology since it ensures secure and collision-free navigation, which is a fundamental aspect of our goals.

References \cite{BAI2018145, 4745874, 6095172} contribute approaches for autonomous navigation and obstacle detection, namely through the integration of sensor data, that are essential for the secure and efficient movement of quadcopters in constrained spaces. These achievements are in perfect line with our project's focus on enabling collision-free coordination among heterogeneous robotic teams.


The versatility and mobility of robotic systems over many terrains are essential for our study. Legged robots, as demonstrated in \cite{saint2023hmas, thakker2021autonomous, li2021openstreetmap}, have notable benefits in navigating uneven terrains filled with obstacles, a capacity that wheeled robots often lack. Reference \cite{4745874} presents ASGUARD, a robot that combines legged and wheeled characteristics to enhance its versatility for urban search and rescue operations. Furthermore, the ability to change the configuration between tracks and wheel legs, as mentioned in \cite{6095172}, increases the mobility of legged robots in emergencies. When combined with the aerial capabilities of drones, as described in \cite{BAI2018145}, it results in significant enhancements in situational awareness and operational efficiency.


Our research utilizes a motion capture system to localize robots in an indoor robotic facility
The review of \cite{6290692}, \cite{7786340}, and \cite{7373476} reveals the strategic utilization of equipment such as Vicon for tracking UAVs with high precision. By utilizing several cameras and identifiers on drones, these systems enable the precise tracking of UAVs' current positions and orientations in real-time. The primary focus in \cite{6290692} is on the utilization of motion capture systems to assist UAV navigation in GPS-denied environments. This study explores the capabilities of these systems that allow drones to follow detailed flight routes and adjust to changes in their operational environments. These technologies provide a solid basis for the development of autonomous UAVs in restricted locations. Furthermore, \cite{7786340} discusses the practical use of safety measures, such as the addition of nylon netting during experimental stages. This method effectively mitigates potential risks connected with UAV testing, such as possible collisions, so assuring the safety of both the technology and the researchers engaged. The studies, \cite{6290692, 7786340, 7373476}, collectively highlight the significant impact that motion capture technology can have in overcoming the difficulties of navigating indoor and other GPS-denied situations. Integrating these technical insights into our exploration improves their operating efficiency and safety in a wide range of challenging environments. This integration expands the possibilities for UAV navigation and control capabilities.

The integration of advanced control systems has been a primary focus in increasing autonomous UAV operations, specifically for completing precise tasks such as autonomous landing. An important development in this field is the utilization of the Robot Operating System (ROS) on a Raspberry Pi 4 to navigate quadcopters and achieve autonomous landing on an ArUco marker \cite{Daspan_2023}. This setup utilizes the Visual Inertial Odometry (VIO) technique for vision-based localization to precisely ascertain the three-dimensional position of the UAV, facilitating accurate navigation and landing on a specified marker. Through 25 landing trials, the experimental results demonstrated that the system has the ability to land with an average precision of 11.12 cm. The integration of ROS with the Raspberry Pi 4 offers a great opportunity to improve the accuracy of UAV operations, which is essential for deployments in GPS-denied conditions such as interior settings or heavily populated locations.

\subsection{Contributions}
The paper contributes a model of safe coordination for a heterogeneous robotic system that consists of a single quadruped robot as the leader and a team of quadcopter robots as the followers (See Fig. \ref{systemoverview0}). 
We treat quadcopters as particles of a $2$-dimensional deformation body and plan safe coordination of the quadcopter team so that  the following objectives are achieved:

\textbf{Objective 1:} The quadruped robot is always enclosed by the convex hull defined by the boundary quadcopters. 

\textbf{Objective 2:} The quadcopter team deformation is characterized such that it can safely pass through narrow passages while following the quadruped robot.

To achieve the above objectives, we first define a Global Coordinate System (GCS), to localize the quadruped and obstacle positions, and a Local Coordinate System (LCS) to safely plan the configurable motion of the quadcopter team. To be more precise, we define the group coordination of the quadcopter team by an affine transformation. The Jacobian matrix of the affine transformation characterizes the rotation and deformation of the quadcopter team formation  with respect to the local coordinate system. Therefore, collision-free coordination of the quadcopter team can be ensured by: (i) polar decomposition of the Jacobian matrix into an orthogonal rotation matrix and a positive definite strain matrix, and (ii) constraining the smallest eigenvalue of the strain matrix.

The proposed model is evaluated by the  \textit{hardware-based} and \textit{mixed virtual-hardware} experiments. For the hardware-based experiment, we localize the environment, quadruped, and quadcopters with respect to the GCS that is fixed on the ground, and evaluate the proposed coordination model in an indoor robotic space. For the mixed virtual-hardware experiment, on the other hand, we localize the quadcopter team and the environment with respect to the local coordinate system that is fixed on the quadruped. Therefore, the environment virtually moves with velocity $-\mathbf{v}$ towards the quadruped that moves with velocity $\mathbf{v}$, with respect to the global coordinate system. Because we measure positions with respect to the local coordinate system,  the quadcopter team purely deforms and rotates without rigid body translation in real flight space  while the configurable motion of the entire robotic team with respect to the global coordinate system is virtually simulated using Gazebo and augmented to the experiment. 

\subsection{Outline}
The paper is organized as follows: The robotic system control is overviewed in Section \ref{System Control} and followed by the modeling approach detailed in Section \ref{approah}. The experimental setup is described in Section \ref{EXPERIMENTAL SETUP}. The results of our experiments are presented in Section \ref{EXPERIMENTAL Results}. Conclusion and plan for the future work are presented in Section \ref{Conclusion}.

\begin{figure}[h]
    \centering
    \includegraphics[width=\linewidth]{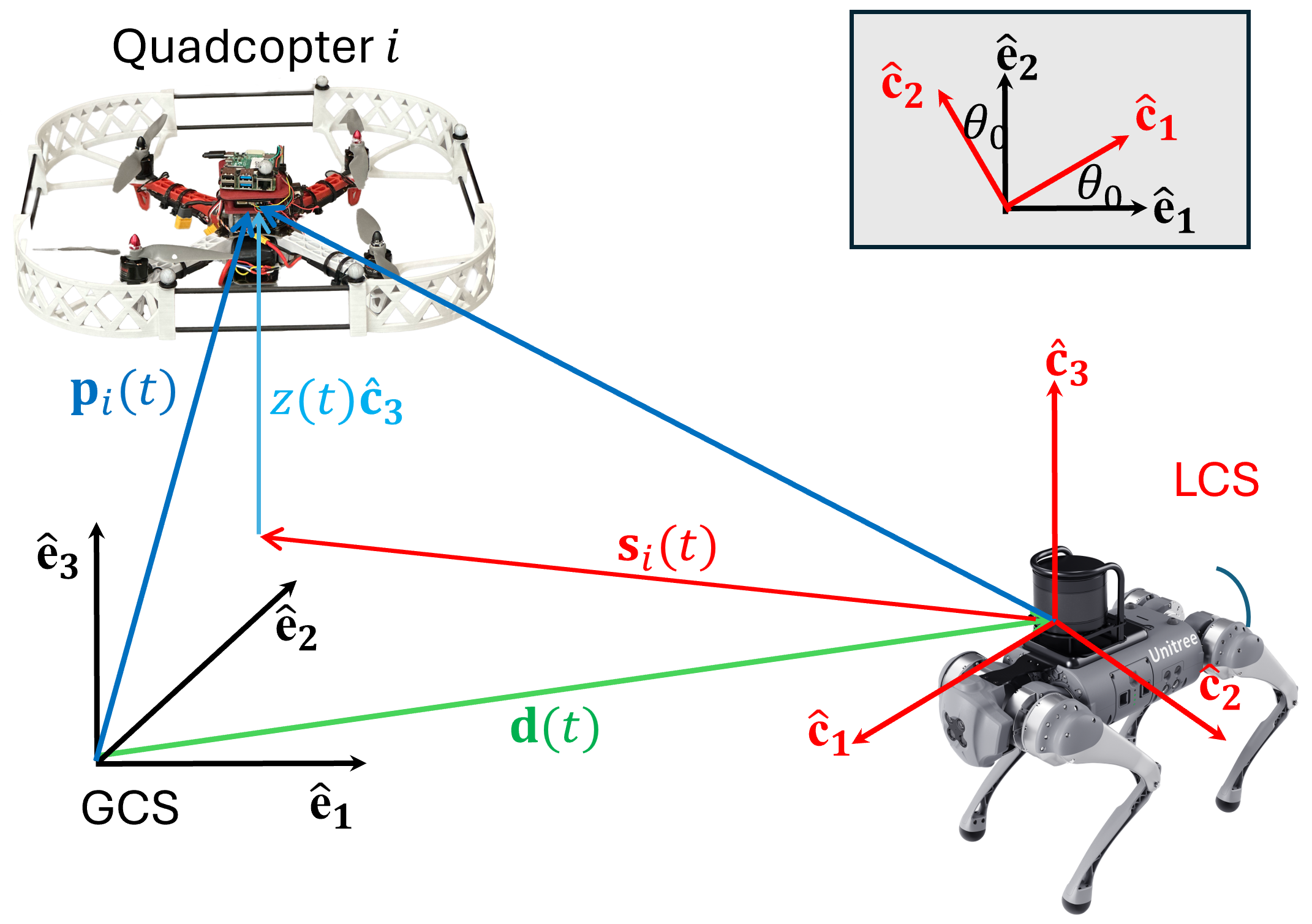}
    \vspace{-0.5cm}
    \caption{Demonstration of Local Coordinate System (LCS) and Global Coordinate System (GCS)}
    \label{LCSGCS}
\end{figure}

\section{System Control}\label{System Control}

The control mechanism of the quadcopter is operated by the open-source software PX4, which follows the standard cascaded control architecture. This architecture utilizes a combination of Proportional (P) and Proportional-Integral-Derivative (PID) controllers.  Specifically, the P controller is in charge of dictating the velocity, while the PID controller is responsible for maintaining velocity stability and  acceleration command.  The maximum allowable horizontal velocity is determined by the parameter \verb|MPC_XY_VEL_MAX|. Furthermore, horizontal and vertical gain parameters can be adjusted using \verb|MPC_XY_P| and \verb|MPC_Z_P|, respectively.

The communication between the flight controller, Pixhawk, and the Raspberry Pi 4, which acts as the companion computer, is essential for off-board control. This link enables a smooth transfer of flight data and control orders, supporting the autonomous capabilities of the UAV. The Raspberry Pi 4 communicates with the Pixhawk using uXRCE-DDS middleware, exchanging crucial information for the operation of the UAV, including navigation commands and telemetry data. The interaction between Pixhawk and Raspberry Pi using the uXRCE-DDS highlights the advanced level of autonomy and control that can be achieved in unmanned aerial vehicle (UAV) systems. Our study shows that by utilizing this communication structure, there is great potential for using UAVs in complex missions that demand high levels of accuracy and independence. These missions can range from delivering specific items to conducting crucial surveillance operations.


\section{APPROACH}
\label{approah}
This work models and experimentally evaluates safe coordination of a team of heterogeneous robots that includes a single quadruped robot and $N$ quadcopters defined by set $\mathcal{V}=\left\{1,\cdots, N\right\}$. 
We assume that the quadcopters are all contained by a triangles contained by boundary quadcopters defined by set $\mathcal{B}=\left\{1,2,3\right\}\subset \mathcal{V}$.

For this purpose, we define a {global coordinate system} (GCS), with the origin and base vectors $\hat{\mathbf{e}}_1$, $\hat{\mathbf{e}}_2$, and $\hat{\mathbf{e}}_3$, that are  fixed on the ground, to localize the position of the quadruped. We use
\vspace{-0.2cm}
\begin{equation}
    \mathbf{d}(t)=d_1(t)\hat{\mathbf{e}}_1+d_2(t)\hat{\mathbf{e}}_2
\end{equation}
to specify the desired position of the quadruped. We also define a local coordinate system (LCS) with the origin and base vectors $\hat{\mathbf{c}}_1$, $\hat{\mathbf{c}}_2$, and $\hat{\mathbf{c}}_3$ that are fixed on the quadruped robot.
 The LCS translates and rotates with respect to the GCS such that $\hat{\mathbf{c}}_3=\hat{\mathbf{e}}_3$ and $\hat{\mathbf{c}}_1$ specifying heading of the dog makes angle $\theta_0(t)$ with $\hat{\mathbf{e}}_1$ (See Fig. \ref{LCSGCS}). Therefore, base vectors of LCS and GCS are related by 
 \vspace{-0.2cm}
\begin{equation}
    \begin{cases}        \hat{\mathbf{c}}_1(t)=\cos\theta_0(t)\hat{\mathbf{e}}_1+\sin\theta_0(t)\hat{\mathbf{e}}_2\\
    \hat{\mathbf{c}}_2(t)=-\sin\theta_0(t)\hat{\mathbf{e}}_1+\cos\theta_0(t)\hat{\mathbf{e}}_2\\
    \hat{\mathbf{c}}_3=\hat{\mathbf{e}}_3\\
    \end{cases}
    .
\end{equation}
The desired configuration of the quadcopter team is a  $2$-dimensional formation in the plane normal to $\hat{\mathbf{c}}_3$ at elevation $z_d(t)$ with respect to the ground. 

\begin{definition}
Quadcopter $1\in \mathcal{B}$ is defined as the primary leader. The local desired position of the primary leader is constant at any time $t$.
\end{definition}

We define reference configuration
\vspace{-0.2cm}
\begin{equation}
    \Omega_0=\left\{\mathbf{s}_{i,0}=u_{i,0}\hat{\mathbf{c}}_1+v_{i,0}\hat{\mathbf{c}}_2:\forall i\in \mathcal{V}\right\}
\end{equation}
and desired configuration
\vspace{-0.2cm}
\begin{equation}
    \Omega_d=\left\{\mathbf{s}_i\left(t\right)=u_{i}(t)\hat{\mathbf{c}}_1+v_{i}(t)\hat{\mathbf{c}}_2:\forall i\in \mathcal{V}\right\}
\end{equation}
for the quadcopter team, where $\mathbf{s}_i(t)$ and $\mathbf{s}_{i,0}$ is related by
\vspace{-0.2cm}
\begin{equation}\label{originalHT}
    \begin{bmatrix}
        u_i(t)\\
        v_i(i)
    \end{bmatrix}
    =\begin{cases}
        \mathbf{s}_{1,0}&i=1\in \mathcal{B}\\
    \mathbf{Q}(t)    \begin{bmatrix}
        u_{i,0}&
        v_{i,0}
    \end{bmatrix}^T&i\in \mathcal{V}\setminus\left\{1\right\}
    \end{cases}
    ,\qquad \forall i\in \mathcal{V}.
\end{equation}
Note that $\mathbf{Q}(t)\in \mathbb{R}^{2\times 2}$ is called \textit{Jacobian matrix}.

   

 \subsection{Planning of Quadcopter Team Deformation}
 In this section, we develop a model for planning of deformation of the quadcopter team by specifying  matrix $\mathbf{Q}(t)$ over the the fixed interval $\left[t_0,t_f\right]$, where $t_0$ and $t_f$ are known \textit{initial} and \textit{final} times. We propose to obtain $\mathbf{Q}(t)$  based on  local positions of the boundary agents $2, 3\in \mathcal{B}$, where the boundary quadcopters' reference positions are given by
 \vspace{-0.2cm}
 \begin{subequations}
 \begin{equation}\label{reference1}
 \mathbf{s}_{1,0}=l_{1,0}\hat{\mathbf{c}}_1,
 \end{equation}
     \begin{equation}\label{reference2,3}
 \mathbf{s}_{i,0}=l_0\left(\cos \theta_{i,0}\hat{\mathbf{c}}_1+\sin \cos \theta_{i,0}\hat{\mathbf{c}}_2\right),~i=\mathcal{B}\setminus \left\{1\right\}.
 \end{equation}
 \end{subequations}
Note that $l_{1,0}$, $l_0$, $\theta_{2,0}$, and $\theta_{3,0}$ are constant. 

 \begin{assumption}\label{assum1}   
 We assume that $\theta_{3,0}>\theta_{2,0}>0$, and $\theta_{i,0}\in \left(0,2\pi\right)$, for $i\in \mathcal{B}\setminus \{1\}$.
 \end{assumption}
Per Assumption \ref{assum1},  boundary agents are not collinear. 
\begin{assumption}\label{assum2}
     We assume that $l_0>0$ and 
     $
 l_{1,0}\geq l_0.
$
\end{assumption}

\begin{figure}[h]
    \centering
    \includegraphics[width=\linewidth]{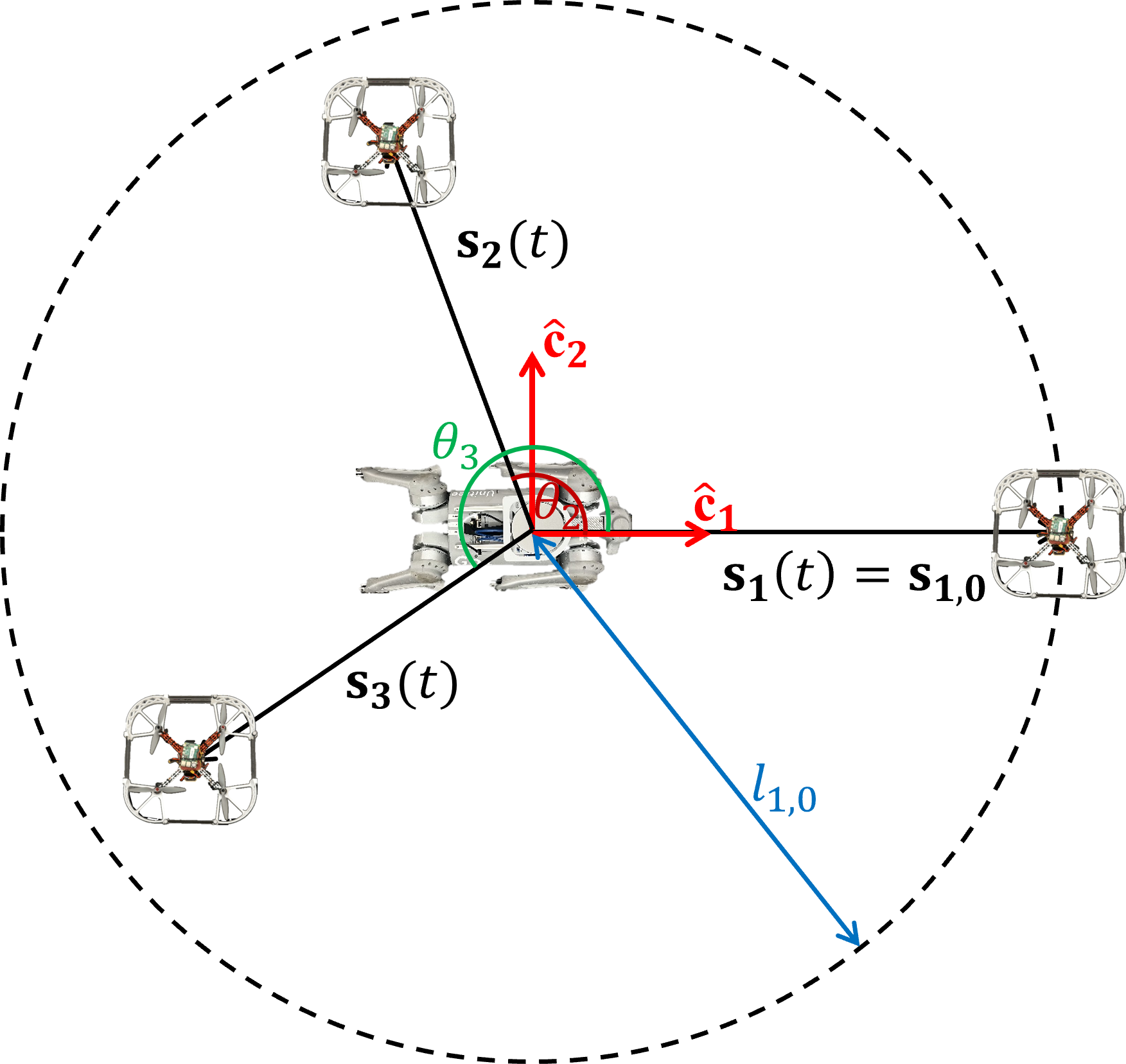}
    \vspace{-0.5cm}
    \caption{Schematic of desired configuration of the robotic system}
    \label{desiredconf}
\end{figure}

Local desired trajectories  of boundary quadcopters $2$ and $3$ are defined by
\vspace{-0.2cm}
\begin{equation}
    \mathbf{s}_i(t)=l_i(t)\left(\cos \theta_i(t)\hat{\mathbf{c}}_1+\sin \theta_i(t)\hat{\mathbf{c}}_2\right),~~ i= \mathcal{B}\setminus \{1\},~t\in \left[t_0,t_f\right],
\end{equation}
where $l_2(t)$, $l_3(t)$, $\theta_2(t)$, $\theta_3(t)$ are considered as the design variables that must satisfy the following constraints:
\vspace{-0.2cm}
\begin{subequations}\label{constraint}
    \begin{equation}
        \bigwedge_{i\in \mathcal{B}\setminus \left\{1\right\}}\left(l_{min}\leq l_i(t)\leq l_0\right),\qquad t\in \left[t_0,t_f\right],
    \end{equation}
    \begin{equation}
        \theta_2(t)<\theta_3(t),\qquad ~t\in \left[t_0,t_f\right],
    \end{equation}
\end{subequations}
where $l_{min}$ is assigned based quadcopter size and  tracking error so that collision between the quadcopters is avoided. The schematic of the desired configuration of the robotic system is shown in Fig. \ref{desiredconf}.

        

\begin{theorem}
If Assumptions \ref{assum1}, then, entries of matrix $\mathbf{Q}(t)$, denoted by $Q_{11}(t)$, $Q_{12}(t)$, $Q_{21}(t)$, and $Q_{22}(t)$, by
\vspace{-0.2cm}
\begin{equation}\label{qplanningequation}
        \begin{bmatrix}
        Q_{11}(t)\\
        Q_{12}(t)\\
         Q_{21}(t)\\
        Q_{22}(t)\\
    \end{bmatrix}
    =
    \mathbf{J}
            \begin{bmatrix}
        l_2(t)\cos\theta_2(t)\\
        l_3(t)\cos\theta_3(t)\\
         l_2(t)\sin\theta_2(t)\\
        l_3(t)\sin\theta_3(t)\\
    \end{bmatrix}
    \end{equation} 
    where 
    \vspace{-0.2cm}
    \begin{equation}\label{Jmatrix}
        \mathbf{J}=\dfrac{1}{l_0\sin\left(\theta_{3,0}-\theta_{2,0}\right)}\left(\mathbf{I}_2\otimes \begin{bmatrix}
        \sin \theta_{3,0}&-\sin\theta_{2,0}\\
        -\cos\theta_{3,0}&\cos\theta_{2,0}
    \end{bmatrix}\right)\in \mathbb{R}^{4\times 4}
    \end{equation}
    is constant, where $\otimes$ is the Kronecker product symbol.

\end{theorem}

%
\textbf{Proof:}
 By substituting $\mathbf{s}_{2,0}$, $\mathbf{s}_{3,0}$, $\mathbf{s}_2(t)$ and $\mathbf{s}_3(t)$ into Eq. \eqref{originalHT}, we obtain the following relation:
 \vspace{-0.2cm}
    \begin{equation}
         \begin{bmatrix}
        l_2(t)\cos\theta_2(t)\\
        l_3(t)\cos\theta_3(t)\\
         l_2(t)\sin\theta_2(t)\\
        l_3(t)\sin\theta_3(t)\\
    \end{bmatrix}
    =\mathbf{H}\begin{bmatrix}
        Q_{11}(t)\\
        Q_{12}(t)\\
         Q_{21}(t)\\
        Q_{22}(t)\\
    \end{bmatrix},
    \end{equation}
    where
    \vspace{-0.2cm}
    \begin{equation}
        \mathbf{H}=l_0\left(\mathbf{I}_2\otimes \begin{bmatrix}
            \cos\theta_{2,0}&\sin \theta_{2,0}\\\cos\theta_{3,0}&\sin \theta_{3,0}\\
        \end{bmatrix}\right).
    \end{equation}
    Because Assumption \ref{assum1} is satisfied, $\sin\left(\theta_{3,0}-\theta_{2,0}\right)\neq 0$ and $\mathbf{J}=\mathbf{H}^{-1}$ exists, and it is obtained by Eq. \eqref{Jmatrix}. Therefore, elements of $\mathbf{Q}(t)$ can be uniquely obtained, based on $l_2(t)$, $l_3(t)$, $\theta_2(t)$, and $\theta_3(t)$, by using Eq.  \eqref{qplanningequation}.

Given $l_{2,f}=l_2(t_f)$, $l_{3,f}=l_3(t_f)$, $\theta_{2,f}=\theta_2(t_f)$, $\theta_{3,f}=\theta_3(t_f)$, the design variable $l_2(t)$, $l_3(t)$, $\theta_2(t)$, and $\theta_3(t)$ are planned by 
\vspace{-0.2cm}
\begin{subequations}
    \begin{equation}\label{lit}
        l_i(t)=l_i(t_0)\left(1-\beta(t,t_0,t_f)\right)+l_i(t_f)\beta(t,t_0,t_f),~~i\in\mathcal{B}\setminus \left\{1\right\},
    \end{equation}
     \begin{equation}\label{thetait}
        \theta_i(t)=\theta_i\left(t_0\right)\left(1-\beta(t,t_0,t_f)\right)+\theta_i\left(t_f\right)\beta(t,t_0,t_f),~~ i\in\mathcal{B}\setminus \left\{1\right\},
    \end{equation}
\end{subequations}
where
\vspace{-0.2cm}
\begin{equation}
\resizebox{0.99\hsize}{!}{%
$
    \beta(t,t_0,t_f)=6\left({t-t_0\over t_f-t_0}\right)^5-15\left({t-t_0\over t_f-t_0}\right)^4+10\left({t-t_0\over t_f-t_0}\right)^3,\qquad t\in \left[t_0,t_f\right],
$
}
\end{equation}
is a strictly increasing function over the time interval $\left[t_0,t_f\right]$. Note that $\beta(t_0,t_0,t_f)=0$, $\beta(t_f,t_0,t_f)=1$, $\dot{\beta}(t_0,t_0,t_f)=0$, $\ddot{\beta}(t_0,t_0,t_f)=0$, $\dot{\beta}(t_f,t_0,t_f)=0$, and $\ddot{\beta}(t_f,t_0,t_f)=0$.




\begin{figure}[h]
    \centering
    \includegraphics[width=\linewidth]{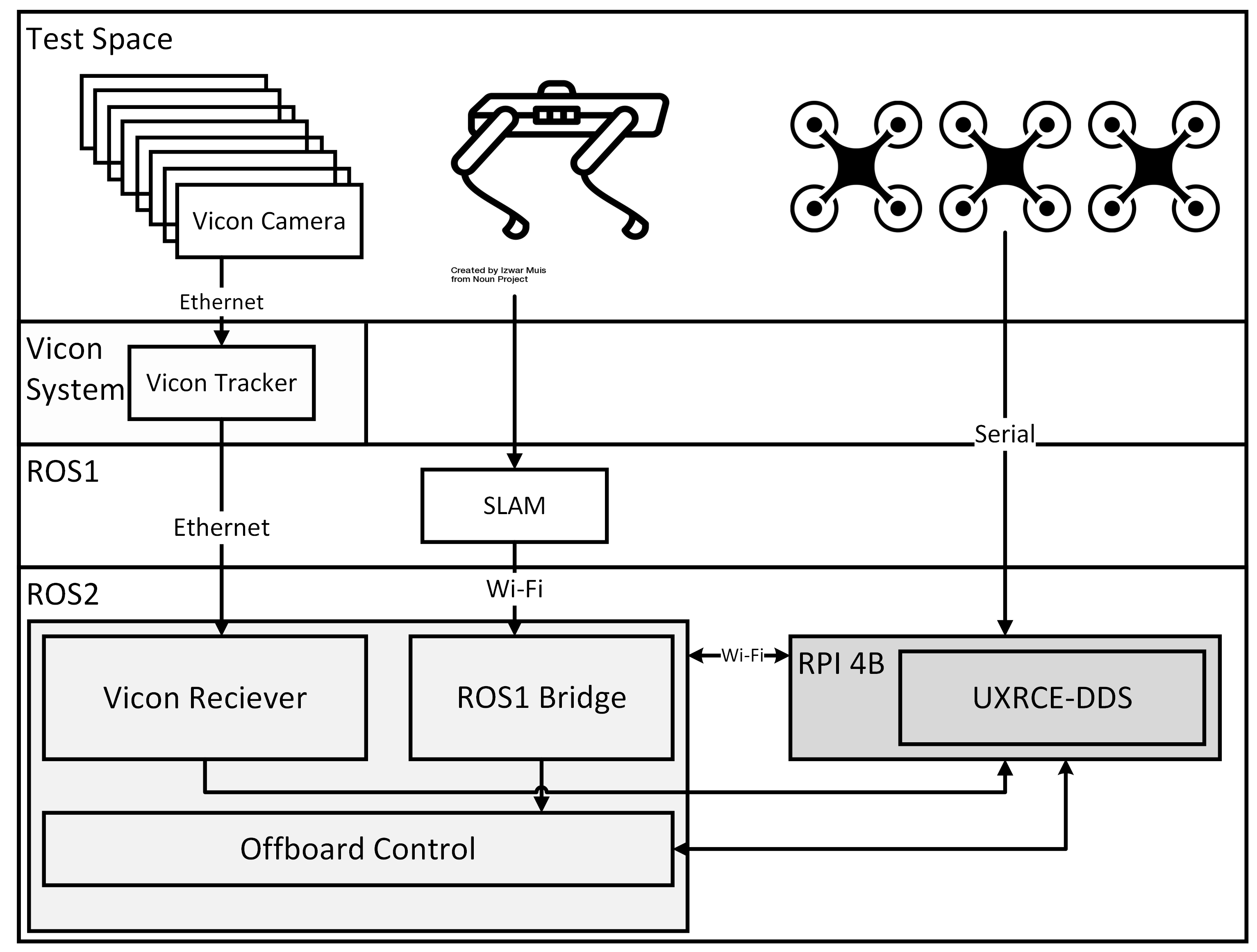}
    \vspace{-0.5cm}
    \caption{Experimental system overview}
    \label{systemoverview}
\end{figure}

\subsection{Safety Assurance}\label{Safety Assurance}
We can ensure collision avoidance between the quadcopters through eigen analysis of Jacobian matrix $\mathbf{Q}(t)$. 
By using polar decomposition, $\mathbf{Q}(t)\in \mathbb{R}^{2\times 2}$ can be decomposed as follows:
\vspace{-0.2cm}
\begin{equation}
    \mathbf{Q}(t)=\mathbf{R}(t)\mathbf{U}(t)
\end{equation}
where
\vspace{-0.2cm}
\begin{equation}
    \mathbf{R}(t)=\begin{bmatrix}
        \cos\psi_r(t)&-\sin\psi_r(t)\\
        \sin\psi_r(t)&\cos\psi_r(t)\\
    \end{bmatrix}
\end{equation}
is an orthogonal rotation matrix, characterized by rotation angle $\psi_r(t)$, and
\vspace{-0.2cm}
\begin{equation}
    \mathbf{U}(t)=\begin{bmatrix}
        \lambda_1\cos^2\psi_d+\lambda_2\sin^2\psi_d&\left(\lambda_1-\lambda_2\right)\cos\psi_d\sin\psi_d\\
        \left(\lambda_1-\lambda_2\right)\cos\psi_d\sin\psi_d&\lambda_1\sin^2\psi_d+\lambda_2\cos^2\psi_d\\
    \end{bmatrix}
\end{equation}
 is a positive definite strain matrix, specifying $2$-D deformation of the quadcopter team, and characterized based on principal eigenvalues $\lambda_1(t)$ and $\lambda_2(t)$ and shear deformation angle $\psi_d(t)$ at any time $t$ \cite{9736156}.
     

 We can ensure inter-agent collision avoidance by constraining the eigenvalues of $\mathbf{U}\left(t\right)$ denoted by $\lambda_1(t)$ and $\lambda_2(t)$ to be greater than $\lambda_{min}$, at any time $t$. Note that the lower bound $\lambda_{min}>0$ is obtained based on the minimum separation distance of all agents, quadcopter tracking error bound, and quadcopter size as detailed in \cite{10156556, 9428540}.

\subsection{Experimental Evaluation Methods}
We propose \textit{hardware-based} and \textit{mixed virtual-hardware} methods  to experimentally evaluate heterogeneous coordination of the quadcopter-quadruped team in an obstacle-laden environment:

\textbf{Method 1-Purely Hardware-Based Experimental Evaluation:} We localize the environment every quadcopter with respect to the global coordinate system by defining global desired  position
\vspace{-0.2cm}
\begin{equation}
    \mathbf{p}_i(t)=\mathbf{d}(t)+z_d(t)\hat{\mathbf{c}}_3+{\mathbf{s}}_i(t)
\end{equation}
and plan $\mathbf{Q}(t)$ such that the quadcopter team safely follow the quadruped robot in a constrained environment.

\textbf{Method 2: Mixed Virtual Hardware Experiment:} We localize the quadcopter team and environment with respect to the local coordinate system, the environment moves towards the quadcopter-quadruped team by relative velocity $\dot{\mathbf{d}}-\dot{\mathbf{r}}$, where $\mathbf{r}=x\hat{\mathbf{e}}_1+y\hat{\mathbf{e}}_2+z\hat{\mathbf{e}}_3$ is position of an arbitrary point of the environment, expressed with respect to the GCS. 

\begin{remark}
Without loss of generality, this paper runs the above experiments by using three quadcopters defined by set $\mathcal{V}=\mathcal{B}$. Therefore, all quadcopters are boundary agents. 
\end{remark}
\section{EXPERIMENTAL SETUP} \label{EXPERIMENTAL SETUP}
We use the available indoor robotic facility in the Scalable Move and Resilient Traversability (SMART) lab to conduct our experiment (the overview of our experiment is illustrated in Fig. \ref{systemoverview}).
 The experimental setup comprises (i) three quadcopters and one quadruped robot, (ii) a motion capture system, and (iii) a ground station computer, as described below.

\subsection{Quadcopter and quadruped robots}
For the experiment, we used three custom-made quadcopters with off-the-shelf components and 3D-printed parts. The F330 frame was selected for the quadcopters due to its adaptability, durability, lightweight design, and affordability. The distance between the two sets of 7-inch propellers, driven by 920kV brushless motors governed by 40A ESCs, is 330mm diagonally. We printed propeller guards to shield them against mild impacts and minimize the damage to the quadcopters and the environment. The quadcopters are equipped with a 4S 2650mAh battery and the PixHawk 6C mini flight controller for its compact size, advanced control features, and dependable software assistance. The quadcopter's flight control, driven by a Pixhawk 4 autopilot, is carefully set up to fulfill the experiment's precise needs. A crucial modification was made to disable sensor fusion capability and utilize the external pose data from Vicon. The control algorithms created in this research use the pose data to dynamically produce waypoints. The waypoints are transmitted to the quadcopter to establish control based on the experimental conditions.

The quadruped robot employs a LIDAR sensor for localization purposes, initially publishing its pose data utilizing ROS1 (Robot Operating System version 1). ROS1's widespread adoption is attributed to its reliability and extensive library of tools and packages that facilitate robot programming. To achieve seamless integration and ensure compatibility with the control system of the quadcopters operating on ROS2, converting the data generated by the quadruped from ROS1 and ROS2 becomes imperative. This strategy of employing the parameter bridge not only enables effective collaboration among diverse robotic platforms but also enhances their operational capabilities. It fosters the development of more sophisticated robotic applications and ensures backward compatibility, allowing systems reliant on ROS1 to interface with the advancements introduced by ROS2 seamlessly.


\subsection{Ground Station Computer (GSC)}
The companion computer transforms the received data into a format compatible with the Robot Operating System (ROS2) and publishes it under the \verb|/fmu/in/vehicle_visual_odometry| topic. The quadcopter uses uXRCE-DDS (eProsima Micro XRCE-DDS) middleware to establish communication between the quadcopter's flight controller module and the companion computer. This middleware facilitates the efficient conversion of uORB messages from the quadcopter's flight control system into ROS2 topics, enabling seamless integration within the ROS ecosystem. The GSC runs the control algorithm and sends the waypoints to each quadcopter using the methodology discussed above.

The GSC also conducts simulations in Gazebo, integrated with the PX4 flight controller, to extend our verification to larger movements constrained by real-world spatial limitations. In these simulations, the quadruped was considered to follow the quadcopters, allowing for a detailed analysis of our coordination model in various scenarios.

\subsection{Motion Capture System}
The SMART Lab, equipped with an eight-camera network, harnesses the high-resolution capabilities of a Vicon motion capture system for tracking precision within a designated area. This system is adept at capturing the pose data of quadcopters by utilizing 5-7 markers on each drone, enabling the accurate determination of their positions and orientations in three-dimensional space. The gathered data is then transmitted directly to a Vicon receiver\cite{ros2vicon_receiver, Garlow2023} at the ground control station (GCS) computer.


\section{EXPERIMENTAL RESULTS}\label{EXPERIMENTAL Results}
We experimentally validated our results by using a quadruped and three quadcopters. The experiments were performed in the Scalable Move and Resilient Transversality (SMART) Lab at the University of Arizona. The facility has an indoor flying area measuring {5}{m} $\times$ {5}{m} $\times$ {2}{m} equipped with $8$ VICON motion capture cameras. To ensure controlled testing conditions, we restricted the quadcopters to a height of {1.5}{m} and confined the movement of the quadruped by {1}{m} $\times ${1}{m}.

For our experiment, we assume that $\theta_2(t)=\theta_{2, 0}$ and $\theta_3(t)=\theta_{3, 0}$, at any time $t$, where we 
choose $\theta_{1,0}=0$, $\theta_{2,0}={2\pi\over 3}$, and $\theta_{3,0}={4\pi\over 3}$. We also choose $l_{1,0}=l_0=1.25m$.  Therefore, Eq. \eqref{qplanningequation} simplifies to
\vspace{-0.2cm}
\begin{equation}\label{qplanningequation}
        \begin{bmatrix}
        Q_{11}\\
        Q_{12}\\
         Q_{21}\\
        Q_{22}\\
    \end{bmatrix}
    ={1\over 1.25}
    \begin{bmatrix}
    0.5000 &   0.5000\\
   -0.2887  &  0.2887\\
   -0.8660   & 0.8660\\
    0.5000    &0.5000\\
    \end{bmatrix}
            \begin{bmatrix}
        l_2(t)\\
        l_3(t)
    \end{bmatrix}
    ,
    \end{equation} 
 where $l_2(t)$  $l_3(t)$  are defined by Eqs. \eqref{lit}.

    \begin{table}[]
        \centering
        \vspace{-0.5cm}
        \caption{Parameters for pure hardware-based experiment}
        \begin{tabular}{|c|c|c|c|c|c|c|}
        \hline
            $t_0$ &$t_f$&$l_{1,0}$&$l_{2,0}$&$l_{3,0}$&$l_{2,f}$&$l_{3,f}$  \\
            \hline
            $0s$ &$35s$&$1.25$&$1.25$&$1.25$&$1.25$&$1.25$  \\
         \hline
        \end{tabular}        
        \label{tab:my_label1}
    \end{table}

\begin{figure}[h]
    \centering
    \includegraphics[width=\linewidth]{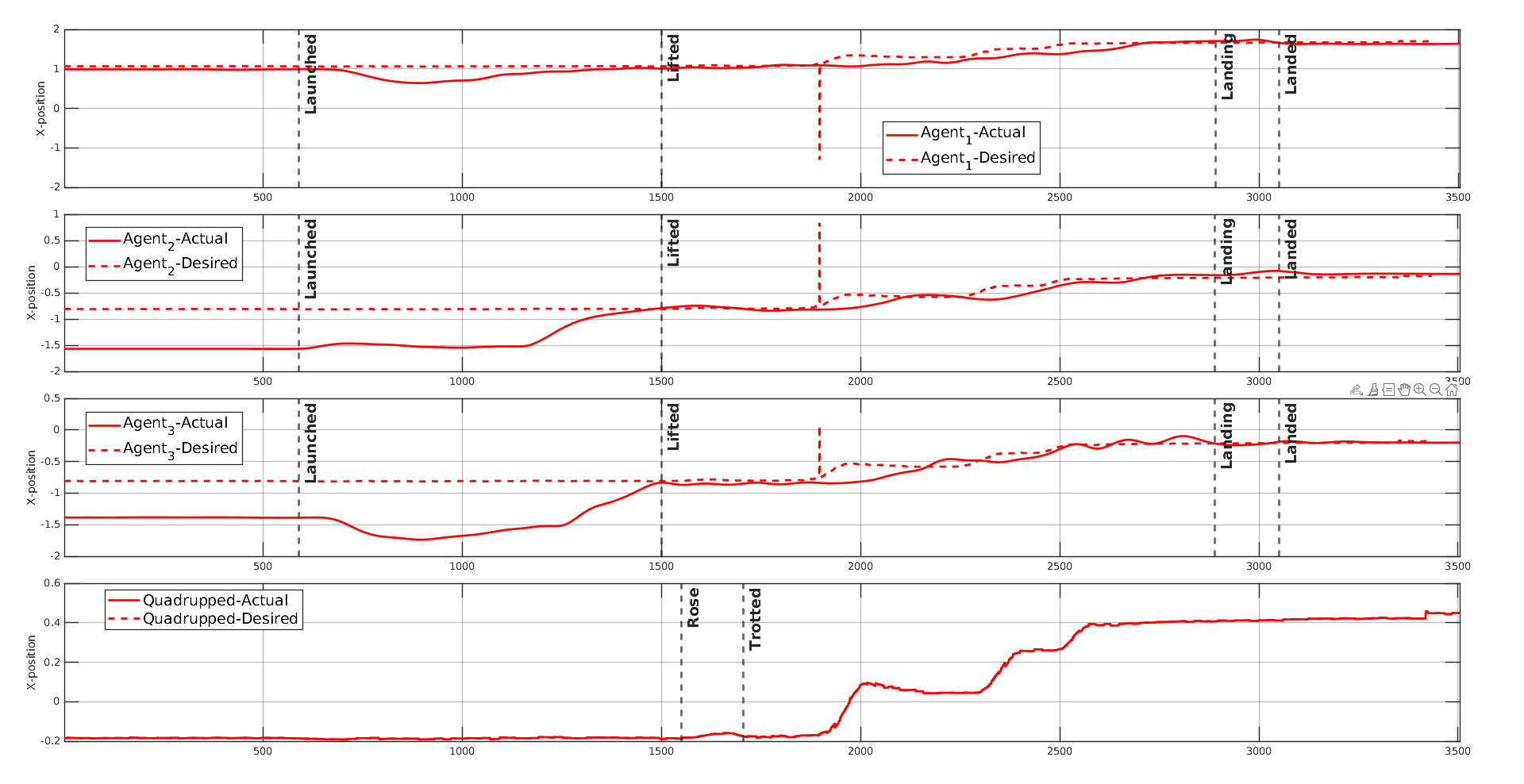}
    \vspace{-0.5cm}
    \caption{Actual versus desired x-position}
    \label{fig:x_plot_exp1}
\end{figure}
\begin{figure}[h]
    \centering
    \includegraphics[width=\linewidth]{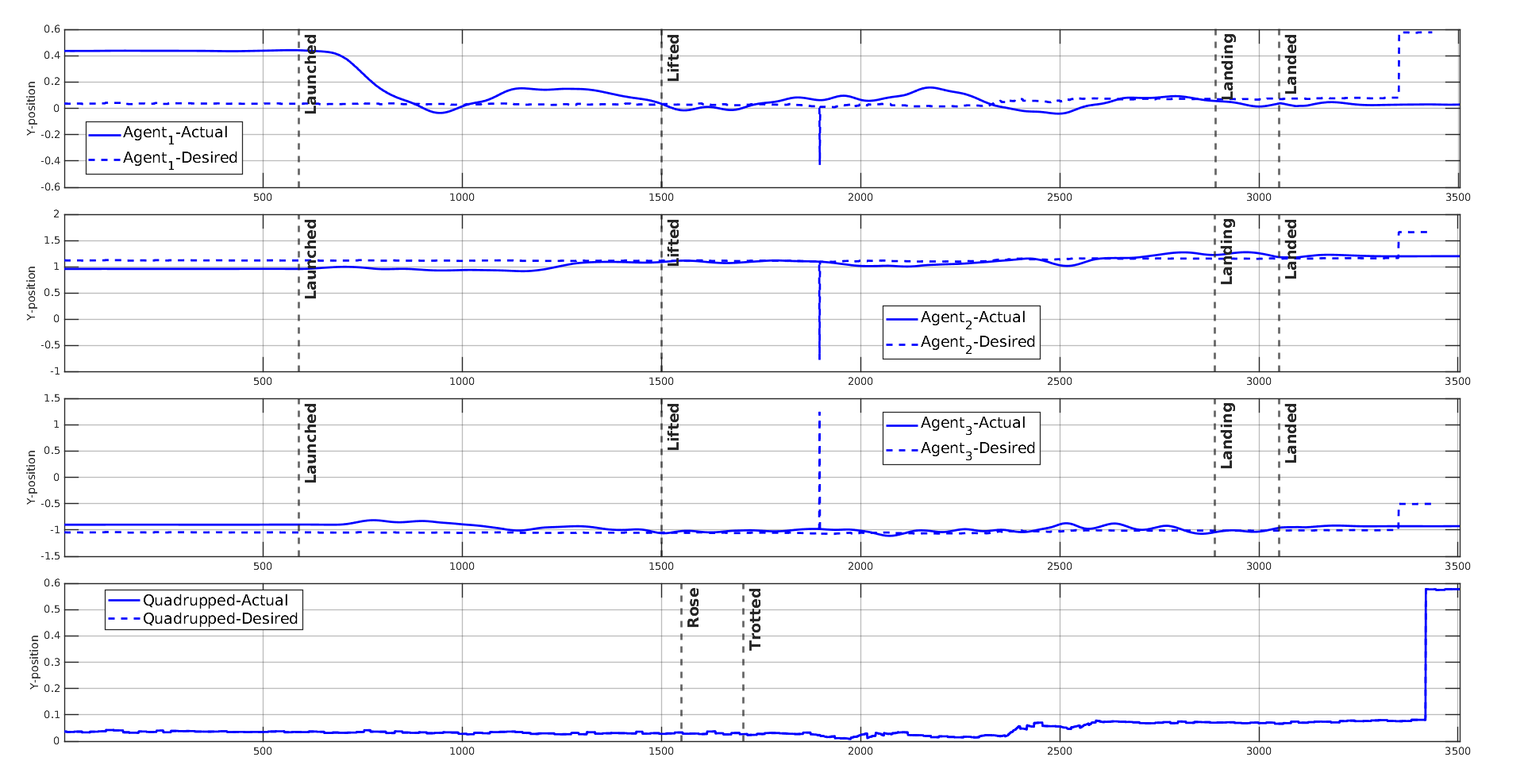}
    \vspace{-0.5cm}
    \caption{Actual  versus  desired y-position}
    \label{fig:y_plot_exp1}
\end{figure}
\begin{figure}[h]
    \centering
    \includegraphics[width=\linewidth]{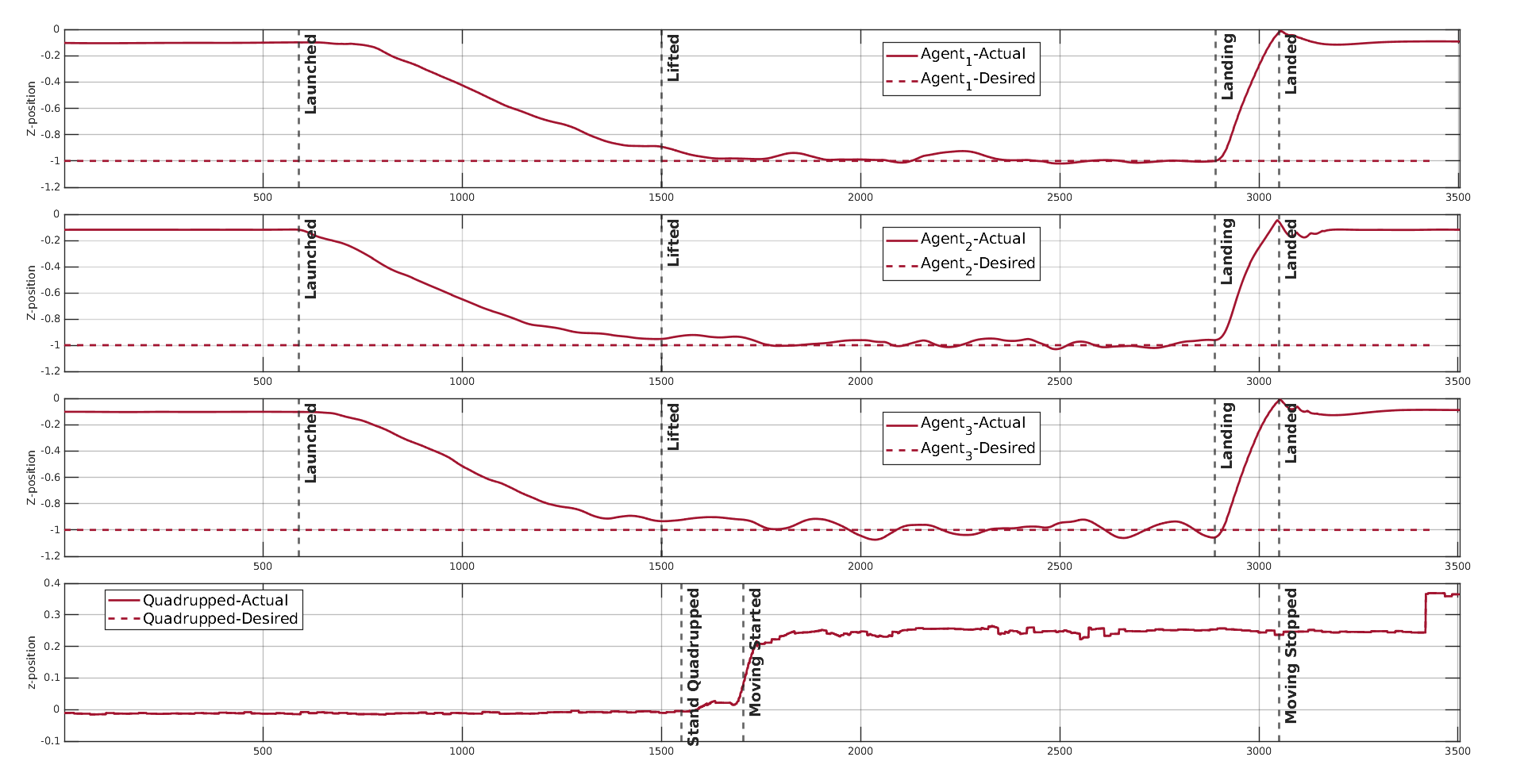}
    \vspace{-0.5cm}
    \caption{Actual  versus  desired z-position}
    \label{fig:z_plot_exp1}
\end{figure}

\subsection{Experiment with Hardware}
For the purely hardware-based experiment, 
the test is conducted by setting $l_{2,f}=l_2\left(t_f\right)=1.25m$ and  $l_{3,f}=l_3\left(t_f\right)=1.25m$ (See Table \ref{tab:my_label1}). For this experiment, the quadruped motion is measured with respect to an inertial coordinate system, fixed at the center of the indoor motion space. The $x$, $y$, and $z$ components of motion of the quadruped are plotted versus time in Figs. \ref{fig:x_plot_exp1}, \ref{fig:y_plot_exp1}, and \ref{fig:z_plot_exp1}, respectively. Note that the $z$ components of the dog changes at sample time $k_s=8500$ when the dog changes its posture from sitting to walking, at the time  it starts its motion.  As shown, the $z$ component of the dog position is almost constant after $k_s=8500$.

By implementing the methodology presented in the \ref{approah}, the quadcopters successfully follow and enclose the dog. Figures \ref{fig:x_plot_exp1}, \ref{fig:y_plot_exp1}, and \ref{fig:z_plot_exp1} also illustrate the components of quadcopters' actual positions, that are shown by continuous plots, and quadcopters' desired positions, that are shown by dashed plots.
\begin{figure}[]
    \centering
    \includegraphics[width=\linewidth]{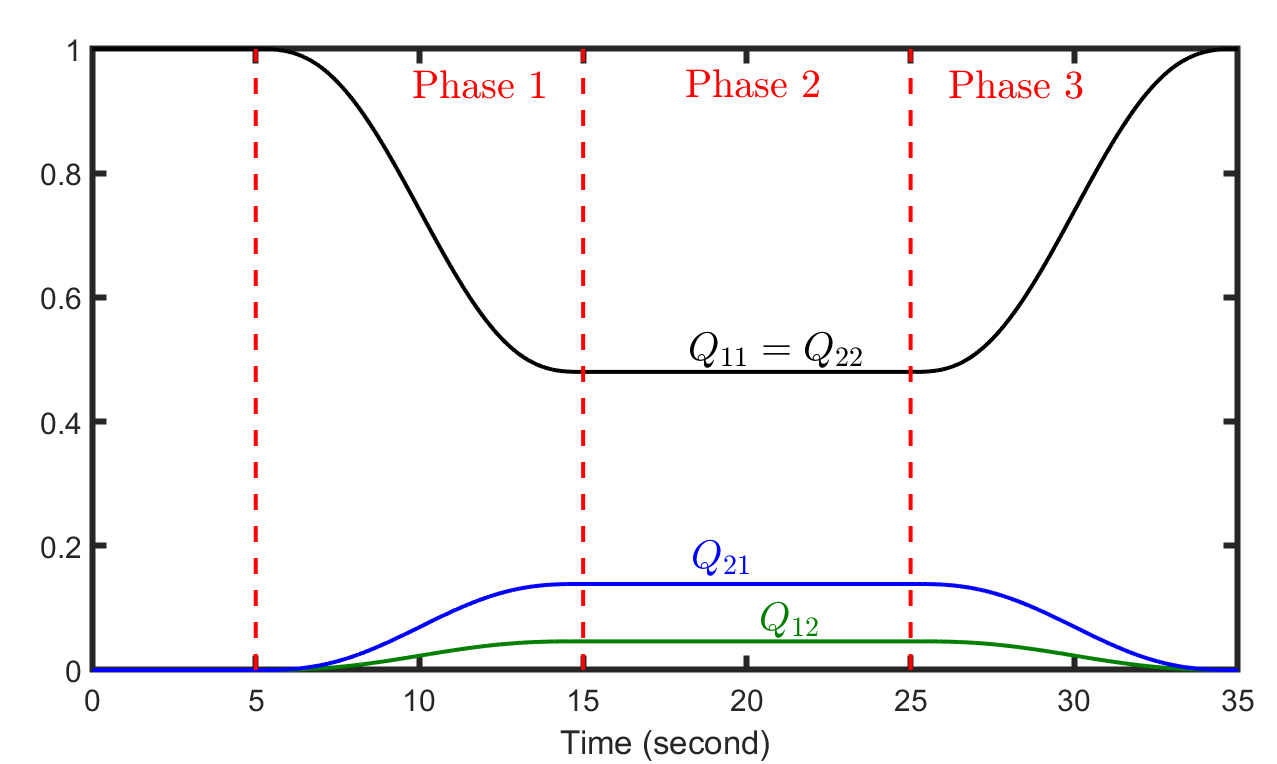}
    \vspace{-0.5cm}
    \caption{Elements of Jacobian matrix $\mathbf{Q}(t)$ versus time $t$}
    \label{Qelements}
\end{figure}
\begin{figure}[]
    \centering
    \includegraphics[width=\linewidth]{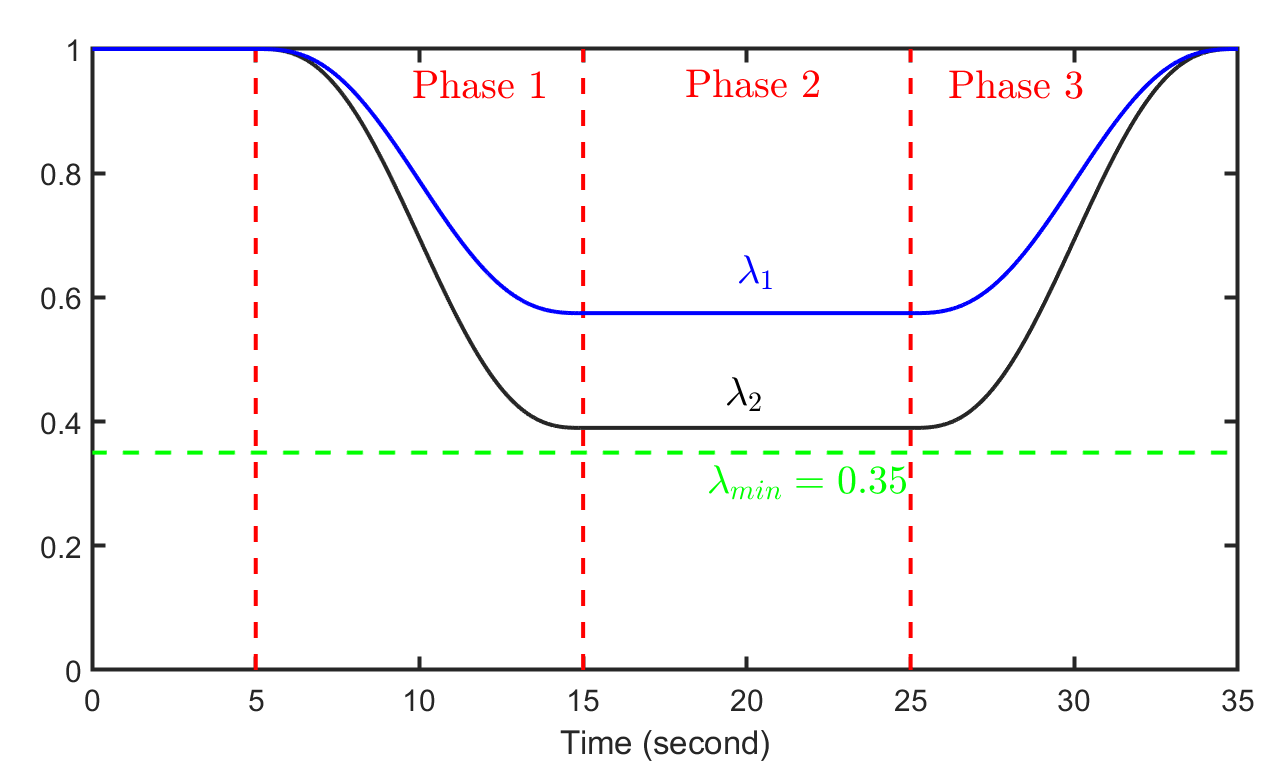}
    \vspace{-0.5cm}
    \caption{Eigenvalues $\lambda_1$ and $\lambda_2$ of  $\mathbf{U}(t)$ versus time $t$}
    \label{Lambdaelements}
\end{figure}
\subsection{Mixed Virtual-Hardware Experiment}
For the mixed virtual-hardware experiment, we consider a scenario at which the robotic system needs to pass through a narrow passage. This scenario is defined as three phases that include contraction (Phase 1), rigid-body translation (phase 2), and expansion (phase 3). We use the parameters listed in Table \ref{tab:my_label2} to set up the experiment. Elements of $\mathbf{Q}(t)$ are obtained by using Eq. \eqref{qplanningequation} and plotted versus time in Fig. \ref{Qelements}. Also, eigenvalues of $\mathbf{U}(t)$ are plotted versus time in Fig. \ref{Lambdaelements}. As shown, $\lambda_1(t)$ and $\lambda_2(t)$ are both greater than $\lambda_{min}=0.35$ to ensure collision avoidance between the quadcopters. 

\begin{figure}[h]
    \centering
    \includegraphics[width=\linewidth]{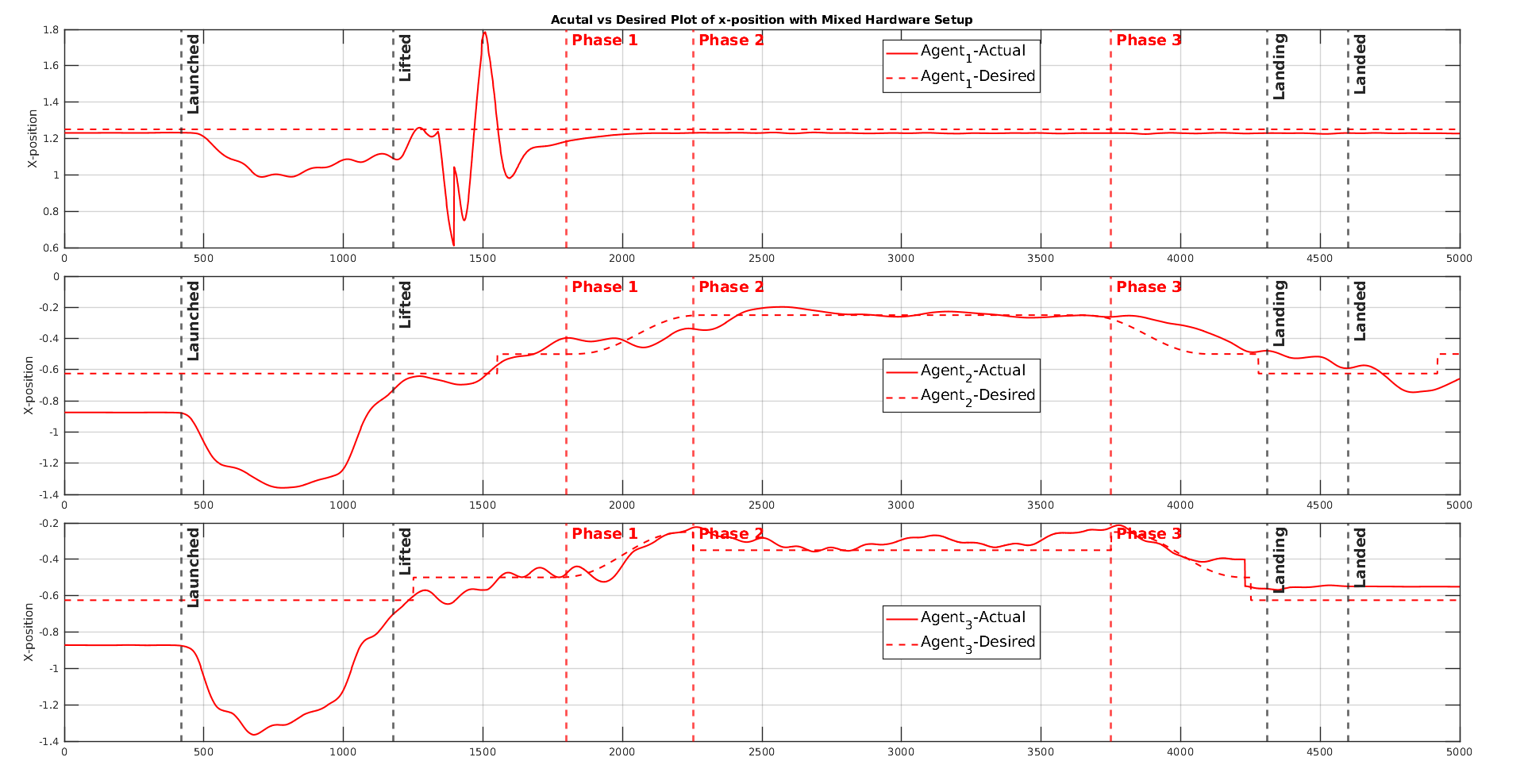}
    \vspace{-0.5cm}
    \caption{Actual versus desired plot of x-position for the mixed virtual-hardware experiment}
    \label{fig:exp_2_x_plot}
\end{figure}

\begin{figure}[H]
    \centering
    \includegraphics[width=\linewidth]{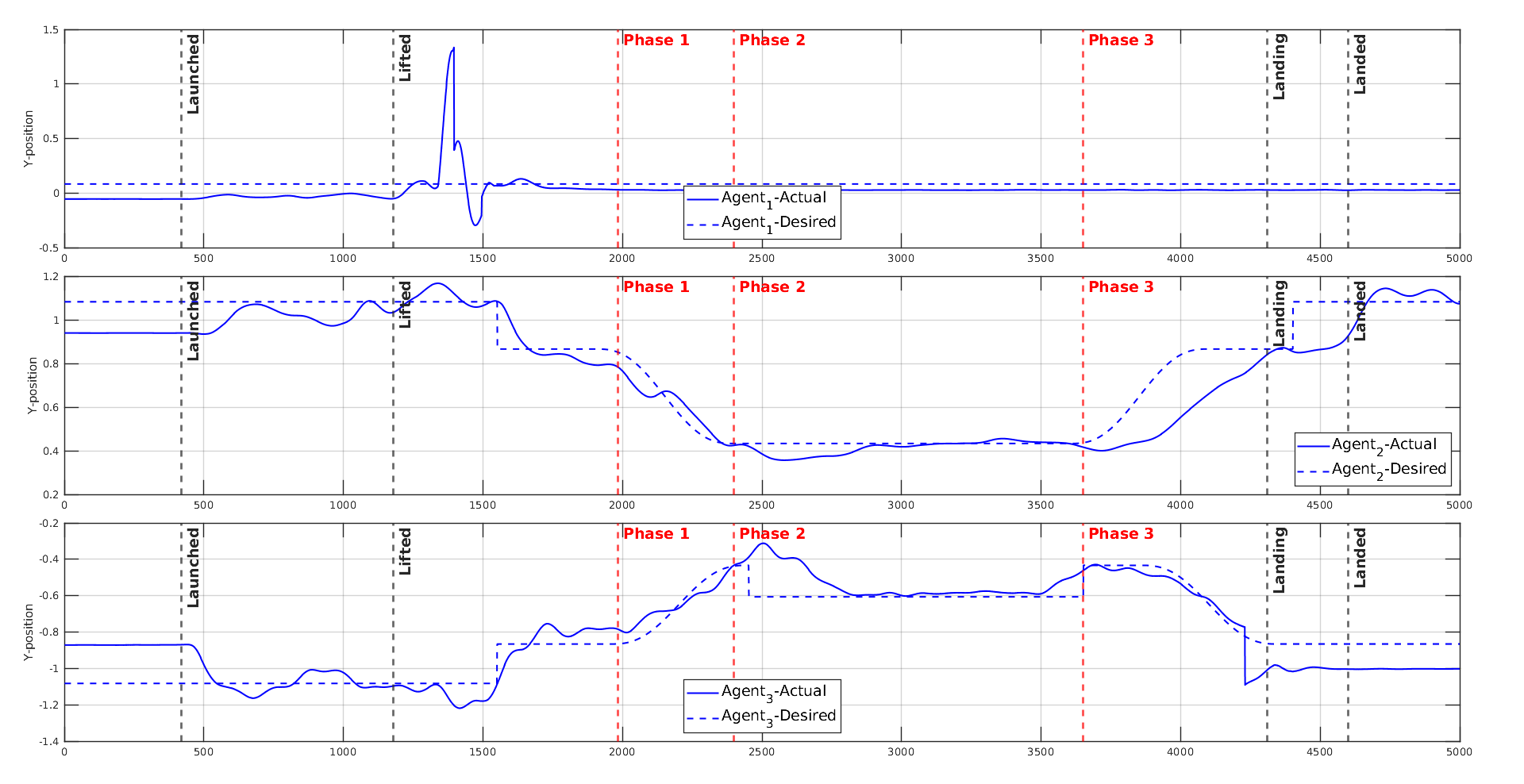}
    \vspace{-0.5cm}
    \caption{Actual versus desired plot of y-position for the mixed virtual-hardware experiment}
    \label{fig:exp_2_y_plot}
\end{figure}

\begin{figure}[h]
    \centering
    \includegraphics[width=\linewidth]{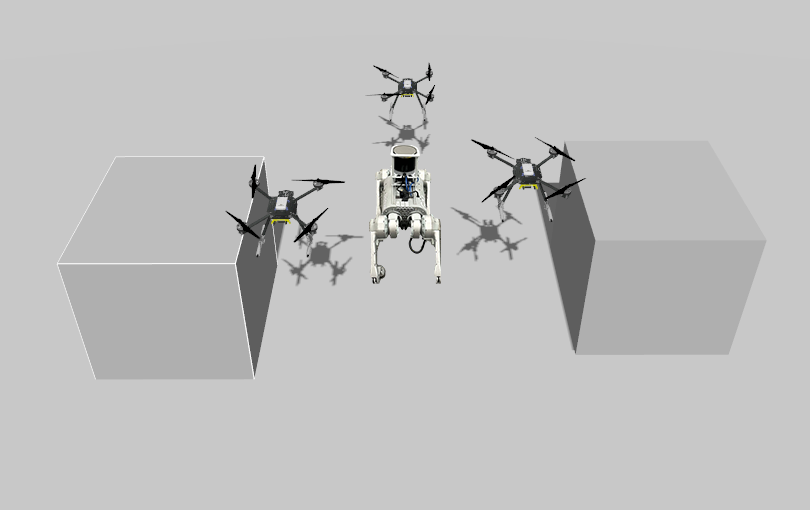}
    \vspace{-0.5cm}
    \caption{A snapshot of the quadcopter formation during phase 2 in the mixed virtual-hardware experiment. Unitree Go1 has been embedded into it.}
    \label{fig:gazebo_image}
\end{figure}

\begin{figure}[ht]
    \centering
    \includegraphics[width=\linewidth]{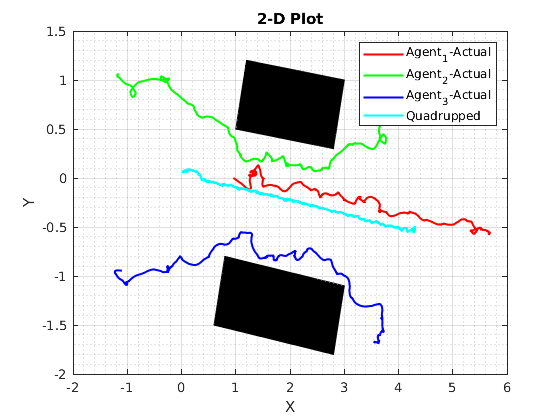}
    \caption{Actual paths of the quadcopters and the quadruped in the mixed virtual-hardware experiment}
    \label{fig:exp3}
\end{figure}

Figures \ref{fig:exp_2_x_plot} and \ref{fig:exp_2_y_plot} illustrate the components of actual and desired local positions of the quadcopters by dashed and continuous plots, respectively. Figure \ref{fig:gazebo_image} shows a snapshot of the quadcopter team formation during phase 2 of this experiment. Also, the actual paths of the quadcopters and the quadruped are shown in Fig. \ref{fig:exp3}.

    \begin{table}[]
        \centering
        \vspace{-0.5cm}
        \caption{Parameters for mixed virtual-hardware experiment}
        \begin{tabular}{|c|c|c|c|c|c|c|c|}
        \hline
             Phases&$t_0$&$t_f$&$l_{1,0}$&$l_{2,0}$&$l_{3,0}$&$l_{2,f}$&$l_{3,f}$  \\
            \hline
           Phase 1& $5s$ &$15s$&$1.25$&$1.25$&$1.25$&$0.5$&$0.7$  \\
           Phase 2& $15s$ &$25s$&$1.25$&$0.5$&$0.7$&$0.5$&$0.7$  \\
           Phase 3& $25s$ &$35s$&$1.25$&$0.5$&$0.7$&$1.25$&$1.25$  \\
         \hline
        \end{tabular}        
        \label{tab:my_label2}
    \end{table}
    






\section{CONCLUSION and Future Work}\label{Conclusion}
This work developed a model for safe coordination of heterogeneous aerial-ground robotic system to navigate through constrained environments. The proposed model defines the desired coordination of the system by an affine transformation which was decomposed into translation mode, specified by the quadruped robot motion, and deformation mode, which was implemented by the team of quadcopters. The proposed approach was successfully tested in an indoor robotic space by using purely hardware-based and mixed virtual-hardware  experiments. For the future, work we plan to experimentally validate configurable motion of a quadcopter team, with both boundary and interior agents, guided by a quadruped robot. Particularly, we will investigate decentralized configurable coordination of the quadcopter team at which the follower quadcopters acquire the desired affine transformation through inter-agent communication with their in-neighbor agents.

\bibliographystyle{IEEEtran}
\bibliography{reference}


\end{document}